\documentclass[letterpaper]{article} % DO NOT CHANGE THIS
\usepackage{aaai2026}  % DO NOT CHANGE THIS
\usepackage{times}  % DO NOT CHANGE THIS
\usepackage{helvet}  % DO NOT CHANGE THIS
\usepackage{courier}  % DO NOT CHANGE THIS
\usepackage[hyphens]{url}  % DO NOT CHANGE THIS
\usepackage{graphicx} % DO NOT CHANGE THIS
\usepackage{natbib}  % DO NOT CHANGE THIS AND DO NOT ADD ANY OPTIONS TO IT
\usepackage{caption} % DO NOT CHANGE THIS AND DO NOT ADD ANY OPTIONS TO IT
\usepackage{algorithm}
\usepackage{algorithmic}
\usepackage{subcaption}   
\usepackage{newfloat}
\usepackage{listings}
\usepackage{amsmath}
\usepackage{booktabs}
\DeclareCaptionStyle{ruled}{labelfont=normalfont,labelsep=colon,strut=off} % DO NOT CHANGE THIS
\floatstyle{ruled}
\newfloat{listing}{tb}{lst}{}
\floatname{listing}{Listing}
\title{SASST: Leveraging Syntax-Aware Chunking and LLMs for Simultaneous Speech Translation}
\author{
  Zeyu Yang\textsuperscript{1}, Lai Wei\textsuperscript{1}, Roman Koshkin\textsuperscript{2}, Xi Chen\textsuperscript{1}, Satoshi Nakamura\textsuperscript{1,3} \\
  \textsuperscript{1}The Chinese University of Hong Kong, Shenzhen, China \\
  \textsuperscript{2}Okinawa Institute of Science and Technology, Japan \\
  \textsuperscript{3}Nara Institute of Science and Technology, Japan  \\
  Correspondence: \texttt{zeyuyang1@link.cuhk.edu.cn}, \texttt{snakamura@cuhk.edu.cn}
}

\begin{document}
\maketitle
\begin{abstract}
This work proposes a grammar-based chunking strategy that segments input streams  into semantically complete units by parsing dependency relations (e.g., noun phrase boundaries, verb-object structures) and punctuation features. The method ensures chunk coherence and minimizes semantic fragmentation. Building on this mechanism, we present \textbf{SASST} (Syntax-Aware Simultaneous Speech Translation), an end-to-end framework integrating frozen Whisper encoder and decoder-only LLM. The unified architecture dynamically outputs translation tokens or \texttt{<WAIT>} symbols to jointly optimize translation timing and content, with target-side reordering addressing word-order divergence. Experiments on CoVoST2 multilingual corpus (En$\rightarrow$De, Zh, Ja) demonstrate significant translation quality improvements across languages and validate the effectiveness of syntactic structures in LLM-driven SimulST systems.
\end{abstract}

\section{Introduction}

Simultaneous speech translation (SimulST) aims to generate target-language translations in real time while listening to ongoing source speech. 
Unlike offline translation, where the entire input is available before translation begins, SimulST must operate under streaming constraints and make decisions dynamically, balancing three often competing goals: translation quality, latency, and output coherence.

Traditional SimulST pipelines typically consist of multiple independent modules, such as automatic speech recognition (ASR), segmentation, and neural machine translation (NMT)~\cite{ma2018stacl,zeng2021realtrans}. 
While modular designs provide flexibility, they also suffer from error propagation, latency accumulation, and a mismatch between training and inference. 
In particular, segmentation and triggering decisions are often based on heuristics or shallow models, lacking deep contextual reasoning and limiting adaptability to varying speech patterns.

Recent progress in large language models (LLMs) has revealed strong abilities in language generation, contextual reasoning, and task generalization~\cite{brown2020language,chowdhery2022palm,openai2023gpt4}. 
This has motivated research into LLM-based SimulST, where powerful sequence modeling capabilities are leveraged for low-latency translation. 
However, existing approaches often retain external policy modules or handcrafted segmentation strategies~\cite{zhang2023diseg}, 
separating ``when to translate'' from ``what to translate'' and thereby limiting interpretability and joint optimization.

In this work, we propose a linguistically motivated, data-driven framework that \textbf{internalizes read/write decision-making into an instruction-tuned LLM}, unifying segmentation and translation within a single model. 
Instead of applying predefined segmentation rules during inference, we generate \textbf{chunk-aligned supervision} based on syntactic and semantic boundaries and use it in a two-stage training strategy to teach the model to predict explicit \texttt{<WAIT>} tokens alongside translation tokens. 
This enables the LLM to autonomously learn when and what to translate, guided by linguistic structure but without relying on external alignment tools or policy heads. 
Inspired by human interpreters, who naturally pause at syntactic or semantic boundaries, our approach yields translations that are more coherent and interpretable under streaming constraints.

To further improve output fluency for language pairs with divergent word orders, we incorporate a chunk-aware reordering mechanism that rearranges translated segments into the natural target-language order. 
Our framework is model-agnostic and can be instantiated with different decoder-only LLM backbones. 
In this work, we evaluate two representative backbones, LLaMA3-8B~\cite{meta2024llama3} and Qwen3-8B~\cite{yang2025qwen3technicalreport}, paired with a frozen Whisper encoder~\cite{cao2012whisper} for speech feature extraction, and operate under causal constraints with all segmentation, alignment, and translation decisions unified within the model itself.

\paragraph{Contributions}
Our main contributions are:
\begin{itemize}
    \item We present a unified, end-to-end SimulST system that integrates translation generation and read/write decision-making into a single LLM.
    \item We propose a linguistically motivated chunk supervision method and a two-stage training strategy that enables the model to autonomously learn translation triggering through explicit \texttt{<WAIT>} token prediction, removing the need for external decision modules or handcrafted rules.
    \item We design a chunk-aware reordering mechanism to improve translation coherence for language pairs with divergent word orders.
\end{itemize}

\section{Related Work}

Simultaneous speech translation (SimulST) aims to deliver translations while speech input is still ongoing, requiring models to balance translation fidelity and latency.  

\subsection{Cascaded Systems}  
Early SimulST systems predominantly adopted a \textbf{cascaded architecture} consisting of automatic speech recognition (ASR) followed by machine translation (MT)~\cite{oda2014optimizing,le2017disentangling}. While effective, these pipelines suffered from error propagation and increased latency due to module coupling. Recent cascaded approaches have leveraged powerful pre-trained models to improve translation quality and latency control. For example, \textbf{BeaverTalk}~\cite{Raffel2025BeaverTalkOS} combines Whisper ASR with an LLM-based translation module. Although these systems reduce latency and improve quality, their multi-module design still inherently couples recognition and translation processes, limiting joint optimization.  

\subsection{End-to-End SimulST}
To overcome the limitations of cascaded architectures, \textbf{end-to-end (E2E) SimulST} models directly map input speech to translations within a single unified neural network, avoiding explicit intermediate transcriptions and module coupling~\cite{berard2016listen,weiss2017sequence,bansal2018pretraining,ren2020simulspeech}. 
By integrating acoustic modeling, language modeling, and translation into a single optimization objective, these systems jointly balance latency and translation quality. 
Early approaches adopted encoder–decoder frameworks with streaming encoders and monotonic attention to handle partial speech input, enabling low-latency generation without waiting for utterance completion. 
Subsequent work further leveraged pretraining, adaptive alignment mechanismss, and multi-task objectives (e.g., simultaneous ASR + translation) to improve robustness and reduce lag. 
Despite their success, these models often require carefully designed read/write policies or specialized attention modules to handle the streaming nature of speech, and their decision-making process for when to emit translations remains either fixed or dependent on external heuristics. 
This limitation has motivated the recent shift towards incorporating large language models (LLMs) into SimulST, aiming to exploit their strong reasoning and generative capabilities while reducing reliance on handcrafted decision modules.

\subsection{LLM-based SimulST}  
Recently, \textbf{large language models (LLMs)} have been introduced into SimulST to exploit their strong reasoning and generation capabilities. 
\textbf{TransLLaMA}~\cite{Koshkin2024TransLLaMA} is one of the earliest works to use LLMs for integrated read/write policy learning, showing that translation triggering decisions can be learned jointly with content generation. 
\textbf{SimulS2S-LLM}~\cite{Deng2025SimulS2SLLMUS} is the first to extend speech LLMs for simultaneous speech-to-speech translation (Simul-S2ST), leveraging boundary-aware speech prompts and a test-time wait-k policy to unlock streaming capability for offline-trained LLMs. 
\textbf{StreamUni}~\cite{guo2025streamuni} further explores unifying segmentation, translation, and generation within a single model using multi-stage reasoning steps. 
These approaches demonstrate the potential of LLM-based architectures for streaming translation but often rely on explicit decision policies or intermediate reasoning stages to determine translation triggers, instead of fully integrating decision-making into the translation process itself.  

\subsection{Read/Write Policies and Segmentation Strategies}  
A key challenge in SimulST is deciding when to emit translation tokens, commonly referred to as the \textbf{read/write policy}. Fixed strategies such as \textbf{wait-k}~\cite{ma2018stacl} and fixed-length chunking~\cite{ma2021streaming} offer predictable latency but lack adaptability. Adaptive approaches learn context-dependent policies through attention analysis~\cite{papi2023attention,Papi2023AlignAttUA}, information-flow estimation~\cite{zhang2022information}, or segmentation-based decision making~\cite{zhang2022adaptive,dong2022learning}. Systems like \textbf{EASiST}~\cite{fu2025efficient} introduce lightweight policy heads, while \textbf{SeqPOS}~\cite{Xu2025SeqPOSiMTSP} frames translation as a sequential decision-making problem using reinforcement learning. Although effective, these approaches often depend on auxiliary classifiers or handcrafted cues, which increase system complexity and limit interpretability.  

Segmentation-based strategies represent another research trend, where models learn to identify translation trigger points at semantically consistent boundaries. Examples include \textbf{RealTranS}~\cite{zeng2021realtrans} with trigger-based decoding, \textbf{MoSST}~\cite{dong2022learning} emphasizing modular SimulST design, and \textbf{DiSeg}~\cite{zhang2023diseg} using differentiable segmentation for improved trigger learning. These methods improve timing interpretability but still treat segmentation as an external or auxiliary process.  

\subsection{Our Approach}  
In contrast, our work adopts a \textbf{linguistically motivated and data-driven perspective}: we identify syntactic and semantic chunk boundaries in bilingual corpora and use them as supervision in a \textbf{two-stage training procedure} to internalize decision-making into the model itself. Rather than relying on separate policy heads, segmentation modules, or heuristic boundary rules, our model jointly produces translations and explicit \texttt{<WAIT>} tokens, learning context-sensitive translation triggers directly within the generation process. This design yields a unified and interpretable SimulST model that integrates boundary reasoning and translation in a single LLM-based architecture, offering a compact alternative to approaches requiring external decision components.

\section{Method}
\label{sec:method}
\subsection{Syntax-Aware Chunking and Chunk-Level Alignment}

A core component of our simultaneous speech translation system is a syntax-aware chunking policy that supervises both read/write decisions and translation timing. 
Unlike fixed windowing or pause-based segmentation methods, 
our approach leverages syntactic information to decide when an input segment is semantically complete and ready for translation. 
This enables the system to produce translation units that align with meaningful linguistic constituents such as clauses and noun phrases, improving semantic focus and fluency under streaming constraints.

To obtain chunk boundaries, we parse source sentences using the \texttt{en\_core\_web\_trf} model from spaCy, which provides token-level part-of-speech tags and dependency relations. 
Chunk segmentation is guided by syntactic boundaries derived from noun phrases (NP), verb phrases (VP), and prepositional phrases (PP), as well as punctuation and dependency transitions (e.g., \texttt{nsubj} $\rightarrow$ \texttt{VERB}). 
Rule-based constraints ensure that each chunk forms a semantically coherent unit and does not exceed a maximum span of seven tokens. 

\begin{figure}[t]
    \centering
    \includegraphics[width=1\linewidth]{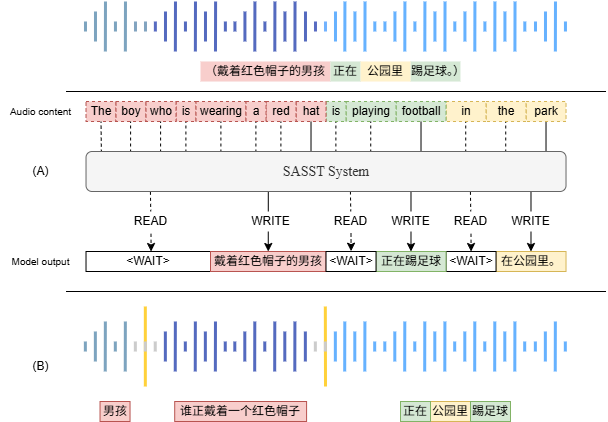}
    \caption{
    Comparison of translation behaviors with syntax-aware training versus pause-based chunking. 
(A) With syntax-aligned training data, our model learns to emit special \texttt{<WAIT>} tokens when encountering incomplete semantic units, 
delaying generation until a complete syntactic chunk is observed, resulting in coherent partial translations. 
(B) In contrast, using a conventional pause-based chunking approach on the same input sentence leads to premature commitments and fragmented outputs, 
highlighting the advantage of syntax-aware chunking in preserving semantic integrity under streaming constraints.}
    \label{fig:alignment}
\end{figure}

To train the model to learn streaming read/write decisions, we construct chunk-aligned bilingual data. 
Given each chunked source utterance, we first obtain fine-grained word-level timestamps using a Whisper-based speech recognizer. 
These timestamps are then aligned to target translations using SimAlign~\cite{jalili2020simalign}, which yields soft bilingual word correspondences. 
For each chunk boundary, the aligned target words are grouped into a translation segment, and a special \texttt{<WAIT>} token is inserted for segments where the model must delay translation. 
This alignment produces training supervision that couples source segmentation with target output timing, enabling causal training of read/write policies without relying on manually annotated delays. 
An example of this chunk-based alignment and its effect on streaming output 
is illustrated in Figure~\ref{fig:alignment}.

\subsection{Target-Side Reordering}
While syntax-aware chunking determines when to start translation, incremental models must also learn what to output when only partial source context is available. 
Directly using the original target sentence can mislead the model because many target words (e.g., verbs in German or function words in Japanese) appear late and depend on unseen source context. 
To address this, we perform a lightweight target-side reordering step to construct training targets that reflect the temporal structure of incremental decoding.

\begin{figure}[t]
    \centering
    \includegraphics[width=1\linewidth]{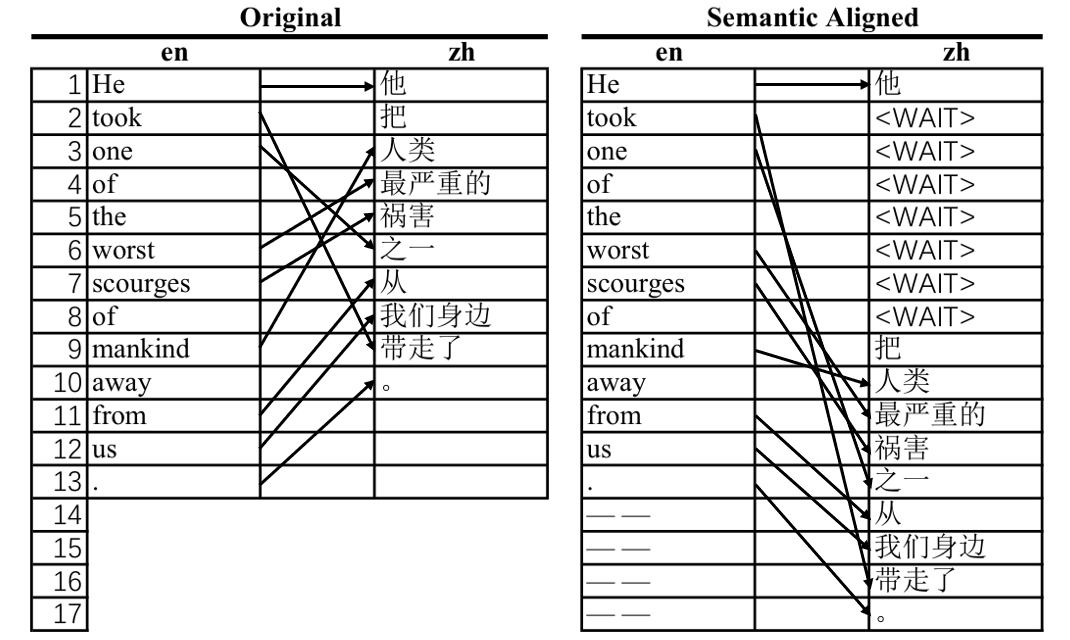}
    \caption{
    Illustration of the target-side reordering strategy.
Target tokens are reordered to match chunk-level source alignment, and special \texttt{<WAIT>} tokens mark positions requiring future context, 
providing explicit supervision for streaming generation timing.}
    \label{fig:reordering}
\end{figure}

Given chunk-level source–target alignments, we rearrange target tokens within each chunk according to their alignment indices and insert special \texttt{<WAIT>} tokens for positions where the model should delay output until more source context arrives. 
This transformation preserves lexical content and grammaticality of the final translation (the reordered target can be deterministically converted back to the original), but it exposes the model to realistic streaming scenarios where partial outputs and waiting decisions are required.
Figure~\ref{fig:reordering} illustrates an example: the original translation (left) places certain arguments and verbs late in the sentence, while our reordered target (right) distributes available words earlier and uses \texttt{<WAIT>} placeholders where future context is necessary.
This supervision allows the model to produce fluent partial outputs while maintaining causal consistency during streaming.

\subsection{System Architecture}
Our system adopts a unified end-to-end architecture that directly maps speech input to translated output under simultaneous translation constraints.
Unlike previous designs that rely on multiple separate modules,
such as independent chunk policy models,
external alignment components, and standalone translation decoders~\cite{oda2014optimizing,ma2018stacl,zeng2021realtrans,bahar2020start}, our approach integrates multiple key functionalities into a single language model backbone. This design allows the model to learn to segment and translate simultaneously within a cohesive generative process, reducing inter-module complexity and improving overall efficiency.

The system consists of a frozen Whisper encoder and a Qwen3-based language model.
Recent work has explored integrating decoder-only LLMs with speech encoders for streaming tasks~\cite{chen2024llast}.
Building on this direction,
our model embeds chunk-aware reasoning into the generation loop, enabling fine-grained control over read/write decisions and unifying segmentation with translation.

\begin{figure*}[t]
    \centering
    \includegraphics[width=0.7\linewidth]{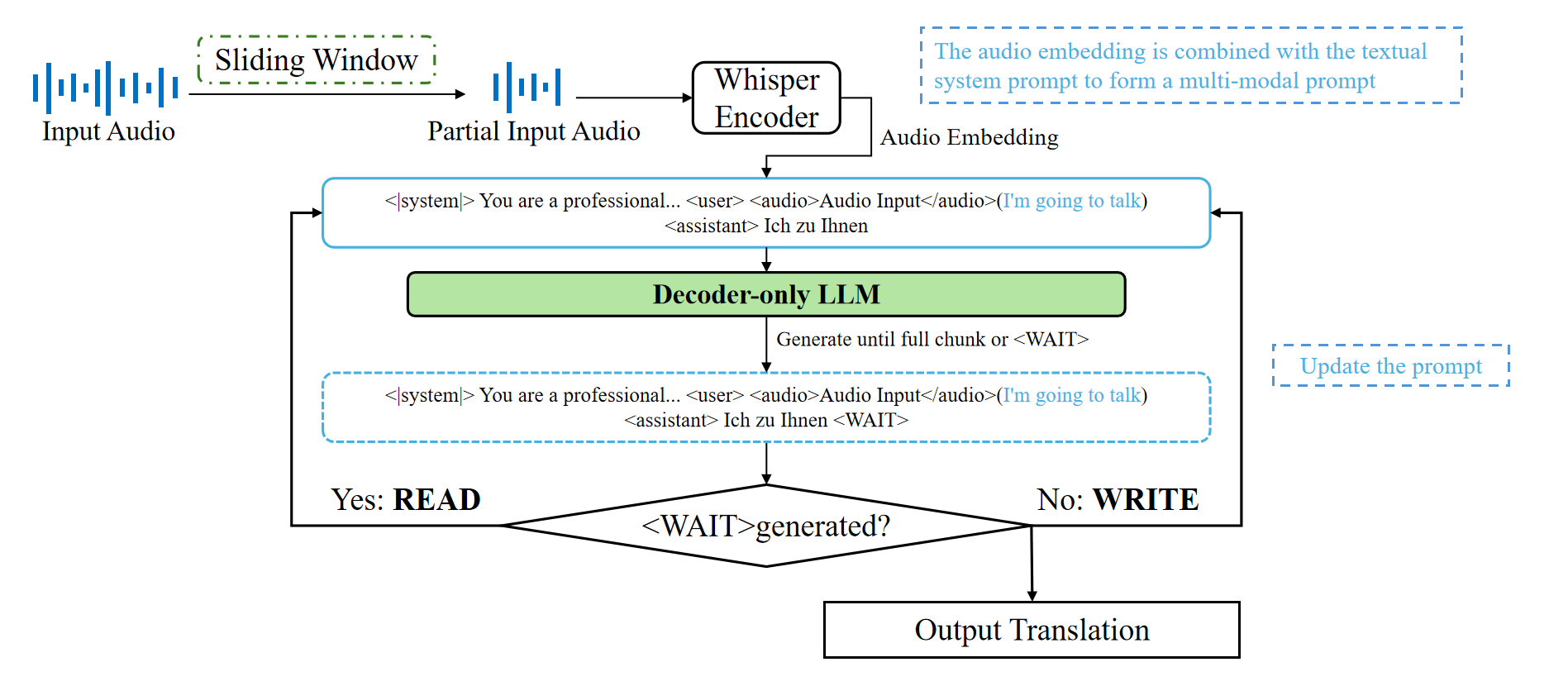}
    \caption{Overview of the SASST architecture for end-to-end simultaneous speech translation.
Input audio is segmented by a sliding window and encoded by a frozen Whisper encoder.
The resulting audio embeddings and textual instruction form a multi-modal prompt for a decoder-only LLM, 
which generates either translation tokens or a special \texttt{<WAIT>} token to control read/write decisions, 
enabling low-latency streaming translation.}
    \label{fig:model-structure}
\end{figure*}

In our design,
chunking and generation are unified into a single autoregressive language modeling task.
The model is trained on streaming sequences,
allowing it to learn natural pause points, maintain coherence over time, and operate under causal decoding constraints. Compared to systems with distinct decision and generation stages, this structure simplifies deployment, reduces error propagation, and enables more effective utilization of large language models for both segmentation and translation. Figure~\ref{fig:model-structure} provides an overview of our architecture, illustrating the interaction between the Whisper encoder, chunk policy mechanism, and autoregressive translation process.

We fine-tune the model on streaming speech data derived from the CoVoST2 corpus~\cite{wang2020covost}, 
a large-scale multilingual speech translation dataset based on the Common Voice project, 
featuring diverse speakers, accents, and spontaneous speech patterns. 
Syntactic chunk boundaries are first extracted from the reference transcriptions using spaCy~\cite{spacy2020} 
and projected back to the source audio via their time-aligned word boundaries, 
yielding a set of audio segments with syntactically informed translation points. 
During training, \texttt{<WAIT>} tokens are inserted between non-aligned regions to supervise timing behavior, 
and the model is trained end-to-end to jointly learn segmentation and translation under causal constraints.

\subsection{Model Training}

We adopt a two-stage training procedure to equip the model with both high-quality 
offline translation capability and read/write decision capability for streaming.

\paragraph{Stage 1: Offline Translation.}
In the first stage, we train the model on full-sentence speech translation pairs 
$\mathcal{D}_\text{offline} = \{(S, Y)\}$, 
where $S = (s_1, \dots, s_T)$ is the input audio waveform 
and $Y = (y_1, \dots, y_N)$ is the full target sentence.
The speech encoder $F_e(\cdot)$, initialized from Whisper, maps input audio to acoustic features 
$H = F_e(S)$. 
A lightweight projection layer $F_p(\cdot)$ converts $H$ into the embedding space of the LLM decoder 
$F_d(\cdot)$ (either LLaMA3-8B or Qwen3-8B). 
We freeze the high-level decoder layers and optionally the speech encoder, 
fine-tuning only the projection and low-level parameters using LoRA adapters~\cite{hu2022lora}. 
The training objective is the standard cross-entropy loss:
\begin{equation}
\mathcal{L}_\text{offline} 
= - \frac{1}{N} \sum_{i=1}^{N} \log P(y_i \mid H, y_{<i}; \theta).
\label{eq:offline_loss}
\end{equation}

\paragraph{Stage 2: Chunk-Aligned Streaming.}
To enable read/write decision learning, we further fine-tune the offline model on chunk-aligned data 
$\mathcal{D}_\text{stream} = \{(S, Y')\}$, 
where the target sequence $Y'$ is augmented with an explicit 
\texttt{<WAIT>} token indicating when the model should pause writing output and continue reading input:
\begin{equation}
Y' = [t_1, \texttt{<WAIT>}, t_2, \dots, t_K].
\end{equation}
The training objective remains the cross-entropy loss but over the extended vocabulary 
including the \texttt{<WAIT>} token:
\begin{equation}
\mathcal{L}_\text{stream}
= - \frac{1}{|Y'|} \sum_{i=1}^{|Y'|}
\log P(y'_i \mid H_{\leq i}, y'_{<i}; \theta).
\label{eq:stream_loss}
\end{equation}
Unlike systems that rely on external segmenters such as SHAS to control read/write behavior, 
our model learns read/write decisions end-to-end through the explicit use of \texttt{<WAIT>} tokens, 
eliminating the need for a separate segmentation module.

\begin{algorithm}[t]
\caption{Syntax-Aware Chunk-Based Streaming Training}
\label{alg:streaming_training}
\begin{algorithmic}[1]
\REQUIRE Chunk-aligned dataset $\mathcal{D}_\text{stream}$, initial parameters $\theta_0$
\FOR{each $(S, Y') \in \mathcal{D}_\text{stream}$}
    \STATE Encode speech: $H = F_p(F_e(S))$
    \FOR{each token position $i$}
        \STATE Predict next token: 
        $\hat{y}_i = F_d(H_{\leq i}, y'_{<i}; \theta)$
        \STATE Compute loss: 
        $\mathcal{L}_i = \text{CE}(\hat{y}_i, y'_i)$
    \ENDFOR
    \STATE Update parameters: 
    $\theta \leftarrow \theta - \eta \nabla_\theta \sum_i \mathcal{L}_i$
\ENDFOR
\end{algorithmic}
\end{algorithm}

\subsection{Streaming Inference and Prompt Encoding}
\label{sec:inference}
\paragraph{1) Simultaneous Inference.}
Our system performs real-time speech translation in a streaming fashion via token-level incremental decoding.
At each step, audio segmented by a sliding window is encoded by a frozen Whisper encoder,
and the resulting embeddings are appended to the source context for the decoder-only language model.
The model decides whether to output a translation token, a special \texttt{<WAIT>} token to defer output, or an \texttt{<EOS>} token to terminate the segment.
During inference, \texttt{<WAIT>} tokens are discarded from the final translation but kept for latency evaluation using SimulEval~\cite{simuleval2020}.

\paragraph{2) Incremental Prompt Encoding.}
Our model adopts a multimodal prompt design inspired by recent LLM-based speech understanding systems.
Each prompt consists of two parts: (1) a fixed instruction text that defines the translation task and streaming behavior,
and (2) a sequence of audio-derived token embeddings extracted from the Whisper encoder.
Unlike prior methods that rely on text transcripts or symbolic prompts, our system operates directly on speech inputs without intermediate ASR, enabling seamless end-to-end streaming translation.

The same multimodal prompt format is used during training and inference,
which reduces domain shift and improves model consistency under streaming constraints.
As decoding progresses,
the prompt is updated incrementally by extending the source-side audio embedding stream and the target-side token history.

This design enables the model to simultaneously reason over speech context,
track translation progress,
and make timing decisions within a unified decoding process.

\paragraph{3) Sliding Window Strategy.}
To support real-time translation while maintaining causal access, 
we apply a sliding window strategy before audio encoding. 
Each input segment is derived from an overlapping 8-second audio window, 
formed by appending the latest $\delta$ seconds of audio to the preceding $8-\delta$ seconds of buffered context. 
This setup preserves both local continuity and long-range acoustic dependencies, 
while avoiding access to future input. 
The stride parameter $\delta$ (e.g., 0.5–2.0 seconds) directly controls the latency–quality tradeoff. 
The Whisper encoder processes each window to extract semantic audio embeddings, 
which are passed to the decoder for joint reasoning.

\section{Experimental Setup}
\subsection{Data}
We conduct experiments on the CoVoST2 dataset~\cite{wang2020covost}, 
which provides speech translation pairs across multiple language directions. 
For this work, we focus on three directions: English$\rightarrow$German (En$\rightarrow$De), 
English$\rightarrow$Chinese (En$\rightarrow$Zh), and English$\rightarrow$Japanese (En$\rightarrow$Ja). 
The CoVoST2 dataset contains approximately 2,900 hours of speech covering 21 languages; 
for our selected pairs, we use the official training, validation, and test splits. 

Following prior work~\cite{dong2022learning,zeng2021realtrans}, 
we evaluate on the official CoVoST2 test set for each language pair. 
In addition, to enable comparison with systems that report results on the ACL 60/60 benchmark, 
we also use this dataset for validation and testing.

\subsection{Evaluation Metrics}
We evaluated the system performance using metrics that capture both translation quality and latency. 
For translation quality, we used BLEU calculated with SacreBLEU~\cite{post-2018-call} 
and COMET~\cite{rei2020comet}. 
For latency, we used the Stream Length-Adaptive Average Lagging (StreamLAAL)~\cite{papi2024streamatt}. 

\subsection{Offline Translation Model}
We train the offline translation model on full-sentence speech–text pairs using the 
AdamW optimizer ($\beta_1=0.9, \beta_2=0.999$) 
with a learning rate of $2.0 \times 10^{-4}$, a warm-up ratio of 0.03, 
and gradient clipping set to 1.0. 
Training is performed for one epoch on 4$\times$V100 GPUs 
with an effective batch size of 32 sentences (16 per GPU with gradient accumulation of 2). 
High-level decoder layers and optionally the speech encoder are frozen, 
while the projection layer and low-level parameters are fine-tuned with LoRA adapters. 
The best checkpoint is selected based on the BLEU score on the development set.

\subsection{Simultaneous Speech Translation Training}
We initialize parameters from the offline translation model and fine-tune on chunk-aligned bilingual data with explicit \texttt{<WAIT>} tokens.
The chunk boundaries and corresponding target-side reordering are derived from our syntax-aware chunking and alignment pipeline described in Section~\ref{sec:method}.
This pipeline produces training examples where each chunk is paired with its aligned translation segment and \texttt{<WAIT>} placeholders for positions requiring delayed output.
Such supervision allows the model to learn when to buffer additional input and when to emit translation in an incremental setting.
During inference, we vary input chunk sizes $\{0.5\mathrm{s}, 0.75\mathrm{s}, 1.0\mathrm{s}, 1.5\mathrm{s}, 2.0\mathrm{s}, 2.5\mathrm{s}, 3.0\mathrm{s}\}$ to explore different quality–latency trade-offs without changing model architecture.

\subsection{Baseline Systems}
We compare SASST with four representative simultaneous speech translation (SimulST) systems, 
all of which leverage large language models (LLMs) and represent the current state of streaming SimulST:

\begin{itemize}
    \item \textbf{BeaverTalk}~\cite{raffel2025beavertalk}: 
    A cascaded pipeline with VAD-based segmentation, Whisper Large V2 ASR, and a LoRA-tuned Gemma~3 12B LLM, supporting both low- and high-latency settings.
    
    \item \textbf{NAIST-2025}~\cite{tan2025naist}: 
    An end-to-end ST model using a Whisper encoder, DeCo projector, and Qwen LLM, with streaming enabled by the Local Agreement (LA) policy and online SHAS segmentation.
    
    \item \textbf{InfiniSST}~\cite{ouyang2025infinisst}: 
    A system for unbounded speech, featuring a chunkwise-causal encoder, speech–text adapter, and multi-turn LLM decoder with KV cache to reduce computation-aware latency.
    
    \item \textbf{SeamlessM4T-IWSLT}\footnote{\url{https://github.com/pe-trik/iwslt25-baselines}}: 
    The official IWSLT~2025 baseline derived from Meta's SeamlessM4T~\cite{barrault2023seamlessm4t}, using a fixed-length segmenter (length 8) to provide stable results across latency regimes.
\end{itemize}

These baselines already cover the most recent streaming LLM-based approaches and thus reflect 
the current practical efficiency–quality trade-offs in the field. 
All models are evaluated on the same \texttt{acl60/60} splits and latency settings for fair comparison.

\section{Main Results}

\subsection{Main Results}

\begin{figure*}[t]
  \centering
  \begin{subfigure}{0.3\linewidth}
    \centering
    \includegraphics[width=\linewidth]{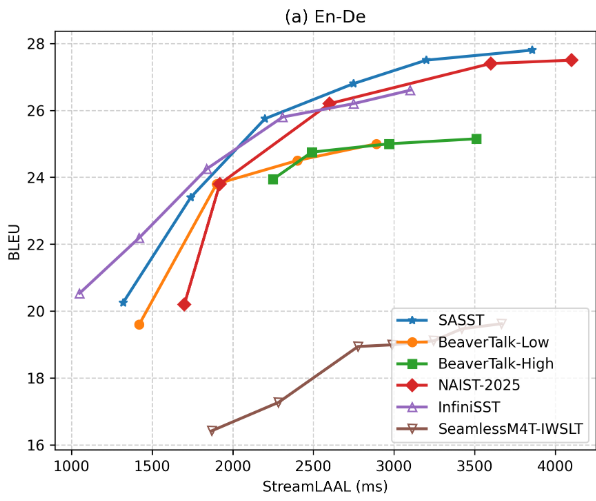}
    \caption{En$\rightarrow$De}
    \label{fig:ende-main}
  \end{subfigure}
  \hfill
  \begin{subfigure}{0.3\linewidth}
    \centering
    \includegraphics[width=\linewidth]{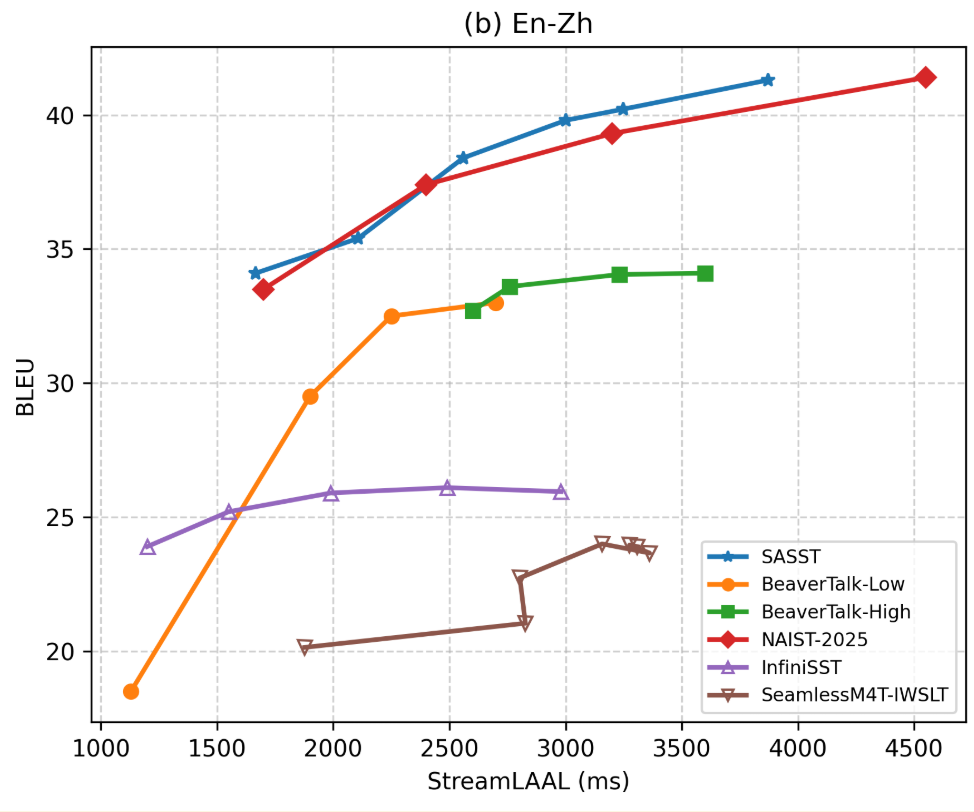}
    \caption{En$\rightarrow$Zh}
    \label{fig:enzh-main}
  \end{subfigure}
  \hfill
  \begin{subfigure}{0.3\linewidth}
    \centering
    \includegraphics[width=\linewidth]{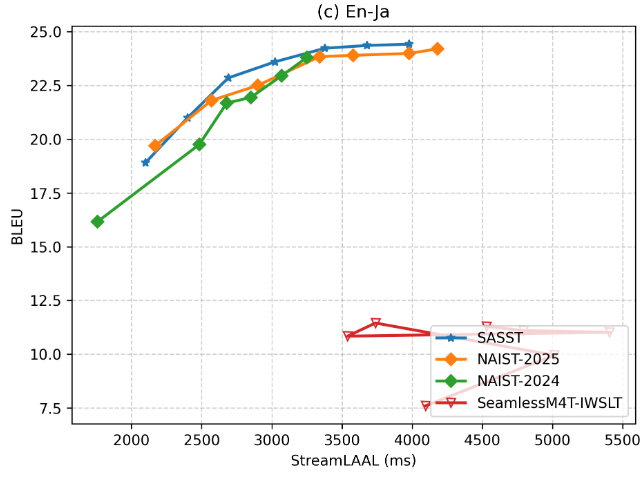}
    \caption{En$\rightarrow$Ja}
    \label{fig:enja-main}
  \end{subfigure}
    \caption{Performance of SASST and IWSLT~2025 baseline systems on 
    acl60/60 En$\rightarrow$De, En$\rightarrow$Zh, and En$\rightarrow$Ja datasets.
    We report BLEU versus StreamLAAL latency to evaluate 
    the quality–latency trade-off for different language pairs 
    with varying syntactic divergence.}
  \label{fig:main-results}
\end{figure*}

Table~\ref{tab:sasst-latency} lists the BLEU scores of our SASST model at 
representative latency points, measured by StreamLAAL.
To better illustrate the latency–quality trade-off, 
Figures~\ref{fig:ende-main}--\ref{fig:enja-main} compare SASST with 
state-of-the-art simultaneous speech translation (SimulST) systems 
from the IWSLT 2025 shared task, including SeamlessM4T-IWSLT, 
BeaverTalk (low- and high-latency configurations), NAIST-2025, and InfiniSST.

\begin{table}[t]
\centering
\small
\begin{tabular}{lcccc}
\toprule
\textbf{Latency (ms)} & 1800 & 2500 & 3200 & 4000 \\
\midrule
\textbf{SASST (En$\rightarrow$De) BLEU}  & 24.6  & 26.2  & 27.7  & 28.0  \\
\textbf{SASST (En$\rightarrow$De) COMET} & 0.729 & 0.744 & 0.758 & 0.762 \\
\midrule
\textbf{SASST (En$\rightarrow$Zh) BLEU}  & 34.1  & 38.5  & 40.2  & 41.5  \\
\textbf{SASST (En$\rightarrow$Zh) COMET} & 0.706 & 0.767 & 0.779 & 0.797 \\
\midrule
\textbf{SASST (En$\rightarrow$Ja) BLEU}  & 18.1  & 22.5 & 23.9  & 24.5  \\
\textbf{SASST (En$\rightarrow$Ja) COMET} & 0.683 & 0.727 & 0.743 & 0.772 \\
\bottomrule
\end{tabular}
\caption{BLEU and COMET scores of SASST at latency levels near 1.8\,s, 2.5\,s, 3.2\,s, and 4.0\,s 
(StreamLAAL), evaluated on CoVoST2 En$\rightarrow$De, En$\rightarrow$Zh, and En$\rightarrow$Ja. 
Exact average latencies may vary slightly due to dynamic chunking.}
\label{tab:sasst-latency}
\end{table}

% ---------------- Figures ----------------

Figure~\ref{fig:main-results} compares SASST  with three representative systems from IWSLT 2025 (the official SeamlessM4T-IWSLT baseline, BeaverTalk, and NAIST-2025) as well as InfiniSST on English–German, English–Chinese, and English–Japanese translation tasks, illustrating the quality–latency trade-offs of different approaches.

Compared to the official IWSLT 2025 baseline, which adopts a fixed-length segmentation strategy and achieves stable but syntax-agnostic performance, SASST demonstrates a superior quality–latency balance across all three language directions within the 2–3.5 second StreamLAAL latency range. For languages such as Chinese and Japanese, which exhibit substantial word-order differences, SASST effectively avoids fragmented outputs caused by insufficient context, maintaining coherent and semantically complete translations. This advantage stems from SASST’s two-stage training on syntax-aligned chunking data, which enables the model to identify semantically complete units and trigger translations at appropriate moments. By internalizing this decision-making capability into the model itself, SASST eliminates the need for external segmentation or triggering modules, thereby reducing error propagation and additional latency, and ultimately achieving more stable and efficient simultaneous speech translation.

\section{Ablation Study}
To deepen the understanding of our approach, we conduct extensive analyses 
under a fixed input chunk size of 2.0\,s to ensure fair comparison. 
We introduce each analytical experiment in detail below.
\subsection{Impact of Syntax-Aware Chunking}
To isolate the effect of syntax-aware chunking, we re-trained SASST using 
a fixed-length segmentation policy and compared it with our syntax-aware 
segmentation on the En$\rightarrow$Zh language pair, which exhibits 
significant word-order differences. 
As shown in Table~\ref{tab:chunking-ablation}, removing syntax-awareness 
causes a substantial drop of more than 15 BLEU points 
(38.5 $\rightarrow$ 23.2) under comparable latency. 
This demonstrates that triggering translations at linguistically meaningful 
boundaries rather than at arbitrary fixed windows is critical for translation 
quality and fluency.

We further analyzed the boundary alignment of the learned chunking policy. 
We measured the proportion of translation triggers that fall within one token 
of a syntactic boundary obtained from an offline dependency parser. 
Our syntax-aware model aligns with syntactic boundaries 82\% of the time, 
compared to only 23\% for fixed-length segmentation, indicating that SASST 
successfully learns to trigger translations near syntactic boundaries, 
resulting in more coherent and semantically complete outputs.

\begin{table}[t]
\centering
\small
\begin{tabular}{lcc}
\toprule
\textbf{Segmentation} & \textbf{BLEU} & \textbf{Boundary Alignment} \\
\midrule
Syntax-aware (Ours) & \textbf{38.5} & \textbf{82\%} \\
Fixed-length        & 23.2          & 23\% \\
\bottomrule
\end{tabular}
\caption{Impact of segmentation strategy on En$\rightarrow$Zh translation. 
Boundary alignment measures the proportion of translation triggers 
aligned with syntactic boundaries.}
\label{tab:chunking-ablation}
\end{table}

\subsection{Influence of LLMs}
We further examine how different foundation models impact SASST performance. 
We compare Qwen3-8B (default) with LLaMA3-8B across three language pairs. 
Table~\ref{tab:llm-ablation} shows that Qwen3-8B consistently 
outperforms LLaMA3-8B by 1.2--3.2 BLEU, while latency remains comparable 
(3.2--4.4\,s StreamLAAL). 
These findings indicate that SASST benefits from the stronger 
instruction-following and multilingual capabilities of Qwen3, 
yet its relative advantage from syntax-aware chunking 
and unified decoding policy is preserved across different LLM backbones.

\begin{table}[t]
\centering
\small
\begin{tabular}{lccc}
\toprule
\textbf{Language Pair} & \textbf{LLM} & \textbf{BLEU} & \textbf{StreamLAAL (ms)} \\
\midrule
En$\rightarrow$Ja & LLaMA3-8B & 25.674 & 4267 \\
                  & Qwen3-8B  & \textbf{27.279} & 4381 \\
En$\rightarrow$Zh & LLaMA3-8B & 37.048 & 3303 \\
                  & Qwen3-8B  & \textbf{40.216} & 3247 \\
En$\rightarrow$De & LLaMA3-8B & 26.684 & 3623 \\
                  & Qwen3-8B  & \textbf{27.892} & 3857 \\
\bottomrule
\end{tabular}
\caption{Impact of LLM backbone on SASST performance.}
\label{tab:llm-ablation}
\end{table}

These results indicate that aligning translation triggers with syntactic boundaries produces more coherent and semantically complete translations without increasing latency. Moreover, SASST maintains consistent gains across different LLM backbones, demonstrating robustness and scalability.
\subsection{Limitations}
While our experiments demonstrate consistent improvements over strong streaming baselines, 
several limitations remain. 
First, our evaluation focuses on three high-resource language pairs on the CoVoST2 benchmark. 
Although these pairs cover typologically diverse structures, future work should investigate 
low-resource and code-switched scenarios to assess cross-domain generality. 
Second, our syntax-aware chunking relies on dependency parsers to determine 
linguistically meaningful translation triggers. 
Although modern parsers achieve over 90\% labeled attachment score (LAS) 
on English, Chinese, Japanese, and German~\cite{dozat2016deep,qi2020stanza}, 
their accuracy may degrade under noisy speech or out-of-domain inputs, 
potentially affecting chunk boundary quality. 
Finally, our chunk-alignment pipeline, while detailed in Section~\ref{sec:method}, 
introduces additional preprocessing steps that may influence downstream performance 
if implemented differently. 
Future work could explore parser-free approaches or joint optimization of segmentation 
and translation to further enhance robustness and portability.

\section{Conclusion}

In this paper, we propose a novel LLM-driven simultaneous
Speech Translation System that allows the LLMs to decide the translation timing and produce output concurrently. Experiments show that SASST delivers competitive translation quality and low latency, indicating strong potential for real-world streaming applications.

\section*{Ethical Considerations}

This work utilizes publicly available large language models (e.g.,
Whisper,
Qwen) for research purposes.
Due to their probabilistic nature, these models may produce inaccurate or biased outputs. All experiments and methods were conducted independently by the authors. We also used ChatGPT to assist with language refinement.

% 参考文献在文档的最后部分
\bibliography{aaai2026}
\end{document}